\documentclass{article}

     \usepackage[preprint, nonatbib]{neurips_2019}

\usepackage[utf8]{inputenc} 
\usepackage[T1]{fontenc}    
\usepackage{hyperref}       
\usepackage{url}            
\usepackage{booktabs}       
\usepackage{amsfonts}       
\usepackage{amsmath}
\usepackage{gensymb}
\usepackage{nicefrac}       
\usepackage{microtype}      
\usepackage[pdftex]{graphicx}
\usepackage{todonotes} 

\title{Principled Training of Neural Networks \\ with Direct Feedback Alignment}

\author{%
  Julien Launay \& Iacopo Poli \\
  LightOn\\ \\
  \texttt{julien@lighton.ai} \\
  \texttt{iacopo@lighton.ai} \\
  \And
  Florent Krzakala \\
  LightOn,   Laboratoire de Physique de l’ENS,\\ Université PSL, CNRS, Sorbonne Université, Université\\ Paris-Diderot, Sorbonne Paris Cité, Paris, France\\
  \texttt{florent.krzakala@ens.fr, florent@lighton.ai} \\
}

\begin{document}

\maketitle

\begin{abstract}
The backpropagation algorithm has long been the canonical training method for neural networks. Modern paradigms are implicitly optimized for it, and numerous guidelines exist to ensure its proper use. Recently, synthetic gradients methods -- where the error gradient is only roughly approximated -- have garnered interest. These methods not only better portray how biological brains are learning, but also open new computational possibilities, such as updating layers asynchronously. Even so, they have failed to scale past simple tasks like MNIST or CIFAR-10. This is in part due to a lack of standards, leading to ill-suited models and practices forbidding such methods from performing to the best of their abilities. In this work, we focus on direct feedback alignment and present a set of best practices justified by observations of the alignment angles. We characterize a bottleneck effect that prevents alignment in narrow layers, and hypothesize it may explain why feedback alignment methods have yet to scale to large convolutional networks.
\end{abstract}

\section{Introduction}
The architectures and optimization methods for neural networks have undergone considerable changes. Yet, the training phase still relies on the backpropagation (BP) algorithm to compute gradients, designed some 30 years ago \cite{rumelhart1986}.

The main pitfalls of BP are its biological implausibility and computational limitations. On the biological side, the weight transport problem \cite{grossberg1987,crick1989} forbids the feedback weights from sharing information with the feedforward ones. On the practical side, as the update of the parameters of a given layer depends on downstream layers, parallelization of the backward pass is impossible. This phenomenon is known as backward locking \cite{jaderberg2016}. Finally, BP prevents the use of non-differentiable operations, even if some workarounds are possible \cite{hinton2012}.
These issues have motivated the development of alternative training algorithms.

\subsection{Related work}
A number of such methods have focused on enhanced biological realism. Boltzmann machine learning \cite{salakhutdinov2009}, Contrastive Hebbian Learning \cite{movellan1991}, and Generalized Recirculation \cite{oreilly1996} all use local learning signals that do not require propagation of a gradient through the network.

Target Propagation \cite{lecun1986,lee2015,bengio2014}, Decoupled Neural Interfaces \cite{jaderberg2016,czarnecki2017}, and Local Error Signals \cite{nokland2019} use trainable modules from which they derive a learning signal. These methods not only enable asynchronous processing of the backward pass, but alleviate some limitations of BP. For instance, local learning signals do not exhibit vanishing gradients, allowing for deeper architectures. They are also inherently regularized and thus less sensitive to over-fitting -- as they don't use a precise gradient on the training set.

\begin{figure}[h]
  \centering
  \includegraphics[width=\linewidth]{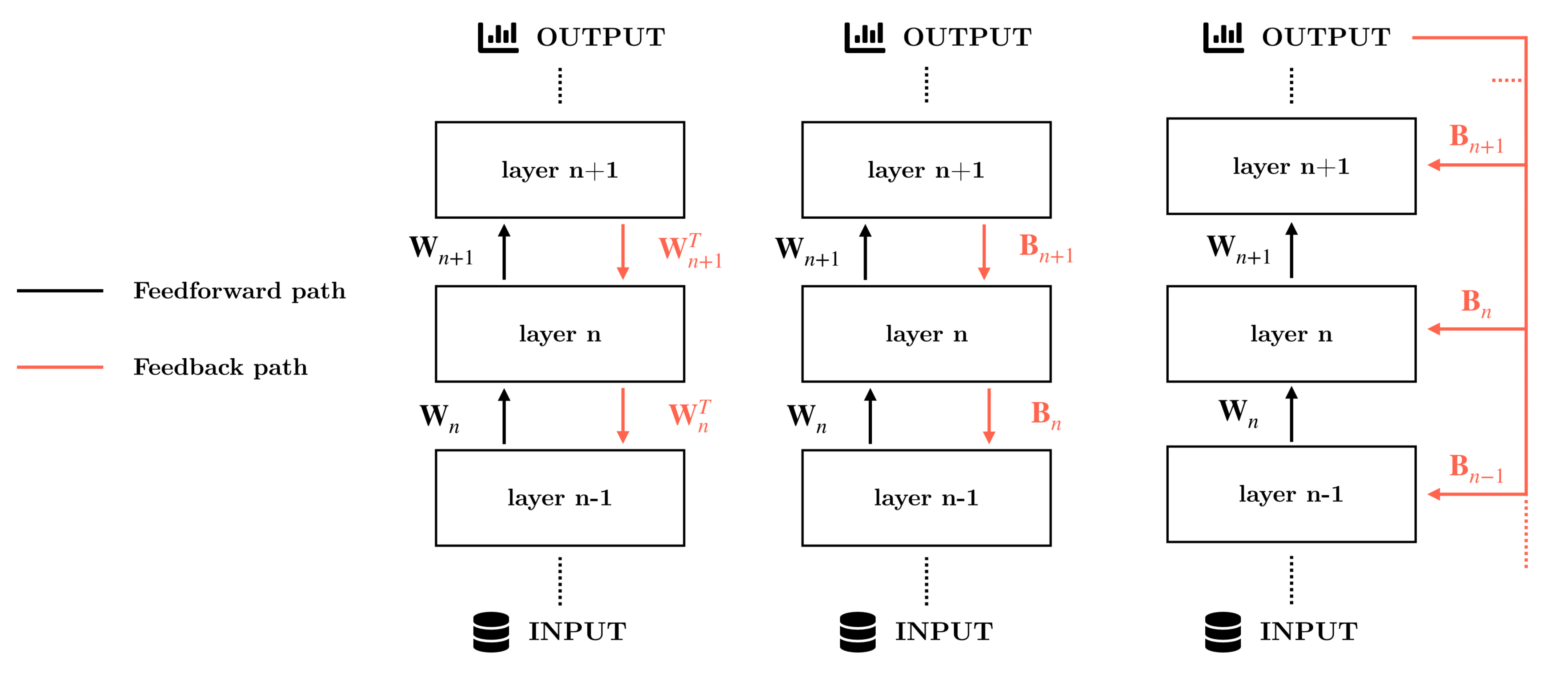}
  \caption{Forward and backward flow in BP, FA, and DFA  (from left to right).}
  \label{fig:bpfadfa}
\end{figure}

Feedback Alignment (FA) \cite{lillicrap2016} leverages the general structure of BP but uses independent feedforward and feedback paths (see Figure \ref{fig:bpfadfa}).  Instead of having the backward weights be the transpose of the forward ones, they are fixed random matrices. Surprisingly, learning still occurs, as the network learns to make the teaching signal useful. This finding has motivated further work around BP with asymmetric feedbacks. If FA comes with a performance penalty, simply keeping a sign-concordant feedback \cite{liao2016} ensures learning with performances on par with BP \cite{xiao2019}. More recently, \cite{akrout2019} introduced a method inspired by alignment where the backward weights are tuned throughout training to improve their agreement with the forward weights, without a direct path between the two to share information. All of these methods solve the weight transport problem, but do not present computational advantages.

Direct Feedback Alignment (DFA) \cite{nokland2016} was introduced as a variant of FA with a direct feedback path from the error to each layer, allowing for layer-wise training (see Figure \ref{fig:bpfadfa}). As this method presents both heightened biological realism and backward unlocking, it is the focus of this paper.

FA and DFA have also been demonstrated with sparse feedback matrices \cite{crafton2019}. However, most of these synthetic gradient methods have yet to scale to harder tasks like ImageNet \cite{bartunov2018}.

\subsection{Our motivations and contributions}
Best practices are well established for training deep neural networks with BP. On the other hand, alternative methods can't rely on such an extensive body of research around their use.

Standard implementations for such methods are scarce. As most deep learning software libraries are designed to perform BP, optimization problems abound when attempting to use them with a different paradigm. Understanding which best practices devised for BP still apply on these different methods can be challenging: quantifying how much of the performance observed is due to the training algorithm and to the architecture chosen is not straightforward. 

Thus, we argue that for DFA and other synthetic gradient methods to be properly evaluated past toy examples, standards must be defined for their use. Only then will we be able to assert their inherent limits. BP did not scale from MNIST to ImageNet immediately; it required years of careful research and considerations. Accordingly, we take the next step towards scaling DFA by presenting a comprehensive set of rules on how to best design and train a deep neural network with this method.

\subparagraph{Alignment angles} Alignment is key to ensuring useful weight updates in DFA. Consequently, properly measuring alignment angles allows us to access novel insights. In Section \ref{sec:method}, we explain our methodology to extend the classic measurement of angles in FA to DFA.

\subparagraph{Baselines} In section \ref{sec:baseline}, we review past claims on best practices for DFA in the light of alignment angles measurements. This allows us to better understand how classic techniques like batch normalization and dropout influence DFA. We also highlight standards for more efficient implementations of DFA, helping alleviate memory issues. 

\subparagraph{Convolutions} In section \ref{sec:conv} we show that convolutional layers systematically fail to align. Thus, DFA is unable to train deep convolutional architectures, a result missed by previous papers. We further hypothesize this may be due to a bottlenecking effect in convolutional layers.

Our implementation of DFA and code to reproduce the results of this paper are available on GitHub\footnote{\url{https://github.com/lightonai/principled-dfa-training}}. 

\section{Method}
\label{sec:method}
 At layer $i$ out of $N$, with $\mathbf{W}_i$ its weight matrix, $\mathbf{b}_i$ its biases, $f_i$ its activation function, and $\mathbf{h}_i$ its activations, the forward pass can be written as:
\begin{equation}
    \forall i \in \left [1, \ldots, N \right ]:\ \mathbf{a}_i = \mathbf{W}_i \mathbf{h}_{i-1} + \mathbf{b}_i,\ \mathbf{h}_i = f_i \left ( \mathbf{a}_i\right )
\end{equation}
$\mathbf{h}_0 = \mathbf{X}$ is the input data and $\mathbf{h}_N = f(\mathbf{a}_N) = \mathbf{\hat{y}}$ are the predictions.

\subsection{Synthetic gradients with Direct Feedback Alignment}
\paragraph{Learning with BP} Parameter updates are propagated through the network using the chain-rule of derivatives, allowing for blame to be assigned precisely to each neuron of each layer. Comparing $\mathbf{\hat{y}}$ and the ground truth $\mathbf{y}$, a loss function $\mathcal{L} = \mathcal{L} \left ( \mathbf{\hat{y}}, \mathbf{y} \right )$ adapted to the task being learned is computed. Disregarding the learning rate, we can write the equation for the update of parameters as: 
\begin{equation}
    \delta \mathbf{W}_i = -\frac{\partial \mathcal{L}}{\partial \mathbf{W}_{i}} = -\left [ \left ( \mathbf{W}_{i+1} \delta \mathbf{a}_{i+1} \right ) \odot f_i'(\mathbf{a}_i) \right ] \mathbf{h}_{i-1}^\top,\ \delta \mathbf{a}_{i} = \frac{\partial \mathcal{L}}{\partial \mathbf{a}_{i}}
\end{equation}
\paragraph{Learning with DFA} In DFA, the gradient signal $\mathbf{W}_{i+1} \delta \mathbf{a}_{i+1}$ coming from the $(i+1)$-th layer is replaced with a random projection of the global error signal. With $\mathbf{e} = \mathbf{\hat{y}} - \mathbf{y}$ the global error vector, and $\mathbf{B}_i$ a fixed random matrix of appropriate shape drawn at initialization for each layer:
\begin{equation}
    \delta \mathbf{W}_i = -\left [ \left ( \mathbf{B}_i \mathbf{e} \right ) \odot f_i'(\mathbf{a}_i) \right ] \mathbf{h}_{i-1}^\top
\end{equation}
The update is thus independent of other layers, enabling parallel processing of the backward pass.
\subsection{Learning and alignment} In BP, the learning signal is defined as:
\begin{equation}\label{eq:ci}
    \mathbf{c}_i = \mathbf{W}_{i+1}^\top \delta \mathbf{a}_{i+1}
\end{equation}
Geometrically, for an arbitrary learning signal $\mathbf{t}_i$ to be useful, it must lie within 90\degree of the signal $\mathbf{c}_i$. This means that, on average, $\mathbf{t}_i^T \mathbf{c}_i > 0$. In the case of feedback alignment:
\begin{equation}\label{eq:dhi}
    \mathbf{t}_i = \mathbf{\delta h}_i = \mathbf{B}_i \mathbf{e}
\end{equation}
Evidently, this means we can ensure learning by tweaking either $\mathbf{B}_i$ or $\mathbf{W}_{i + 1}$. In DFA, as $\mathbf{B}_i$ is fixed, this means the feedforward weights $\mathbf{W}_{i + 1}$ will learn to make the teaching signal useful by bringing it in alignment with the ideal BP teaching signal. This process is key to allowing learning in asymmetric feedback methods.

\paragraph{Measuring alignment} This alignment phenomenon can be quantified as the angle between the DFA and BP signals. We can thus compute the alignment angle $\beta_i$:
\begin{equation}
    \cos(\beta_i) = \frac{\mathbf{\delta h}_i^\top \mathbf{c}_i}{\lvert \lvert \mathbf{\delta h}_i\rvert \rvert \lvert \lvert \mathbf{c}_i\rvert \rvert}
\end{equation}

Accordingly, if $\cos(\beta_i) >0$, then $\lvert \beta_i \rvert < 90^\circ$, and we have learning occurring with DFA. 

From a practical point of view, it is therefore possible to measure alignment angles in DFA, only at the expense of having to perform a standard BP backward pass as well. In the common case of mini-batch gradient descent, the alignment angle is the mean of the diagonal of the matrix resulting from the product of the DFA and BP updates. This is interesting, because by quantifying how certain practices affect the alignment angle, we can better separate which part of the final change in performance observed is due to the practice itself, or to its interaction with DFA.

While alignment angles can be calculated on any parts of the network, we recommend measuring them after the non-linearities of each layer. For simple tasks where very low training errors can be reached, such as MNIST, it is common for alignment towards the end of training to decrease and spread: this is because as the error gets small, so does the DFA training signal. Accordingly, a small change in the distribution of the weights will be enough to generate large change in alignment.

\subsection{Implementation and infrastructure details}

Our implementation of DFA leverages PyTorch \cite{paszke2017automatic}. For all the experiments of the paper, memory requirements were around 1 GB; even consumer-grade hardware should have no problem reproducing our results. We ran our experiments on a machine with 4 RTX 2080 Ti, allowing us to run multiple experiments in parallel.

For larger architectures, our code is able to store the feedback weights and perform the random projections on a separate GPU. Similarly, for angle measurements, the model running backpropagation can be stored on a separate GPU as well. Combined with using a unique feedback matrix (see Section \ref{sec:unique} for details) this allows us to scale DFA to networks with an unprecedented number of parameters. In experiments unreported in this paper, we were able to train a model of more than 217 million parameters on ImageNet with DFA using two V100s. 

\section{Establishing best practices for DFA}
\label{sec:baseline}

Past works have made a number of statements regarding best practices for feedback alignment methods. However, most of these claims have been derived simply from intuition or the observation of loss values. This makes it challenging to discern between the influence of the recommended practices on DFA itself, and the learning capacity they add to the network as a whole. Using angle measurements, we formulate principled recommendations for training neural networks with DFA. We also present some implementation details to make DFA leaner in terms of memory usage.

\subsection{Enabling DFA to scale to very large networks by using a unique feedback matrix}
\label{sec:unique}

The computational advantage of backward unlocking in DFA comes at the price of an increased memory demand. Large random matrices need to be stored in memory to compute the feedback signals. For each layer, the corresponding projection matrix is of size $e \times l_i$ where $e$ is the length of the error vector (e.g. the number of classes in a classification problem), and $l_i$ is the output size of layer $i$. Increasing the output size of the network by a factor $k$ has little effect on the memory needs of BP, while it will scale the size of each feedback matrix by $k$ in DFA, with drastic effects on the memory requirements.

\begin{figure}[h]
  \centering
  \includegraphics[width=0.6\linewidth]{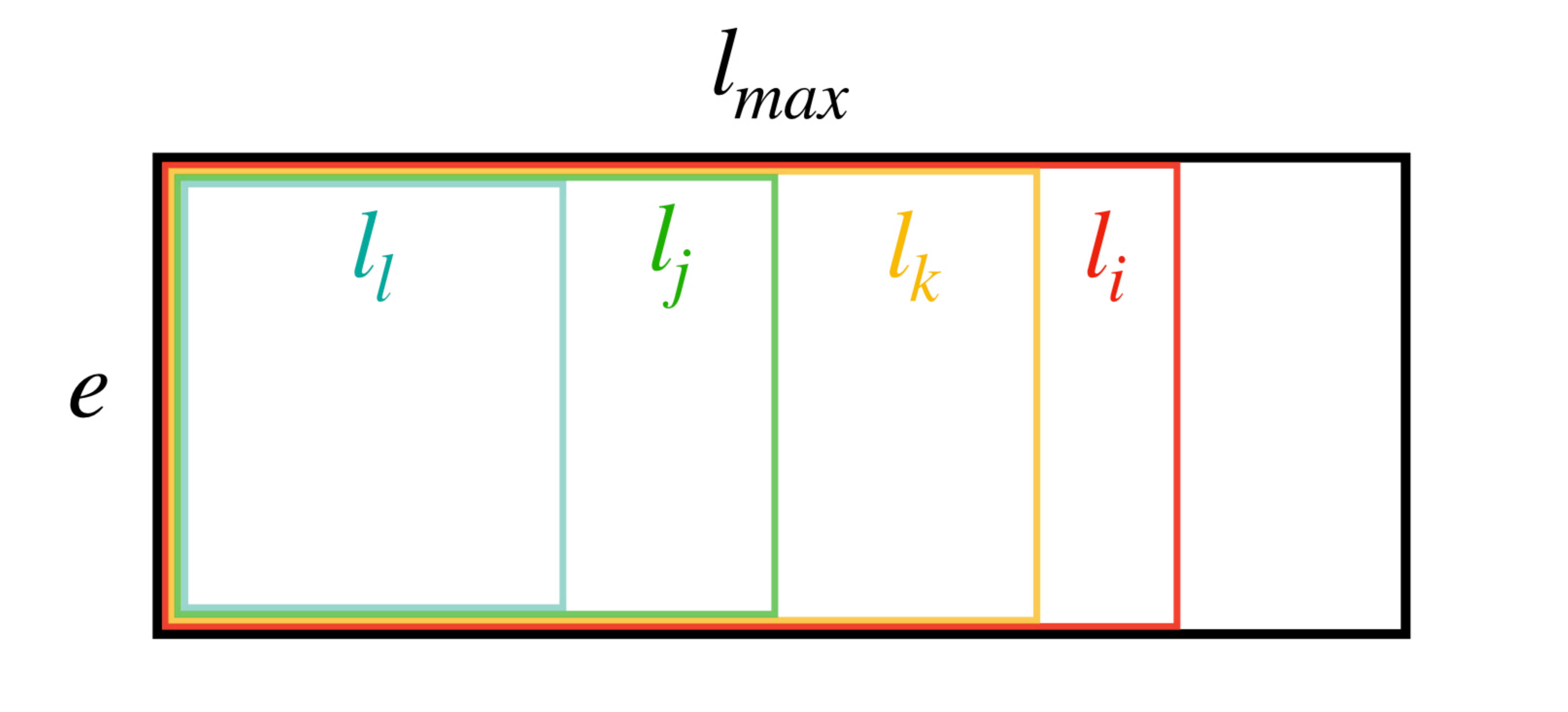}
  \caption{From the feedback matrix of the largest layer with output size  $l_\text{max}$, the feedback matrices of all others layers (here $l_i$, $l_k$, $l_j$, $l_l$) can be obtained by slicing.}
  \label{fig:slicing}
\end{figure}

Thus, a naive implementation of DFA will incur a tremendous memory cost. This has prevented researchers in the past from scaling their experiments to datasets like ImageNet \cite{bartunov2018}. 

We notice that in the theorem justifying DFA convergence \cite{nokland2016} no assumption is made on the independence of the random matrices. In fact, it is even suggested to use the same matrices for layers of the same size. We expand on this idea and implement the backward pass of DFA by drawing just one random matrix for the largest layer. For the smaller layers we take fixed slices of this unique larger random matrix (see Figure \ref{fig:slicing}). We thus reduce the memory requirements, especially for very deep networks, as the memory needed for the synthetic gradients computation will depend only on the size of the largest layer and of the error vector. Table \ref{tab:memory} showcases the savings in memory achieved using our technique. In section \ref{sec:conv}, this allows us to use DFA on ImageNet to train VGG-16.

\begin{table}[h]
\centering
\caption{Comparison of memory cost of the feedback weights in the backward pass between a naive implementation of DFA and our unified feedback matrix implementation. Assuming a network with $N$ layers of output sizes $l_i$ for the i-th layer and an error vector of length $e$. $l_\text{max}$ is the output size of the largest layer. VGG-16 architecture for ImageNet taken from \cite{simonyan2014very}.}
\label{tab:memory}
\begin{tabular}{lcc}
\toprule
 & \textbf{Memory complexity} & \textbf{VGG-16 [GB]}\\
\cmidrule(r){2-3}
naive & $\sum_{i = 1}^{N - 1} l_i e$ & 55\\
unified feedback matrix & $l_\text{max} e$ & $13$\\
\bottomrule
\end{tabular}
\end{table}

\subsection{Revisiting usual best practices in the light of alignment angles}

\paragraph{Normalization of the backward weights} The importance of normalization for the backward weights in asymmetric feedback methods is well-known. However, practices diverge, from methods depending on the feed-forward weights \cite{xiao2019}, to manually tuned parameters \cite{lillicrap2016}. Using information on the weights of the forward path goes against the idea of fixing the weight transport problem. Instead, we propose to simply scale the feedback weights depending on their dimensions, in a way that is compatible with the use of a unique feedback matrix. 

With $\mathbf{U}$ a random matrix such that $\mathbf{U}_{ij} \sim \mathcal{N}(0, 1)$, $e$ the length of the error vector, and $l_\text{max}$ the output size of the largest layer, our unique random matrix $\mathbf{B}$ is defined as:
\begin{equation}
    \mathbf{B} = \frac{\mathbf{U}}{\sqrt{l_\text{max} e}}
\end{equation}

Then, we define for each layer $\mathbf{B}_i \in \mathbb{R}^{l_i \times e}$ by slicing $\mathbf{B}$ to the appropriate dimensions $\mathbf{B}_{1:l_i,:}$, where $l_i$ is the output size of the $i$-th layer. The subscript $1:l_i,:$ means that we take only the first $l_i$ rows and all the columns of the matrix. We adjust the normalization accordingly:
\begin{equation}
    \mathbf{B_i} = \sqrt{\frac{l_\text{max}}{l_i}}\mathbf{B}_{1:l_i,1:e}
\end{equation}

Table \ref{tab:norm} and \ref{tab:conv} show that this normalization significantly improves accuracy and alignment angles, in both fully-connected and convolutional architectures.

\begin{table}[t]
\centering
\caption{Performance of a 3 layers of 800 neurons network on CIFAR-10 with different practices. Learning rate is set to $5 \cdot 10^{-4}$ for SGD with no momentum, except for $\lvert x \rvert$ and lReLU(-0.5) where it is $10^{-4}$. Weights are initialized using He initialization \cite{he2015}. If unspecified, non-linearities are tanh. When dropout is used, it is with probability 0.1 on the input, and the reported value in parentheses for each layer but the output. Accuracies reported are the average of 10 random runs. Statistics for alignment angles are computed on a batch of 128 samples from the best run of the 10. Values in parentheses are standard deviations.}
\label{tab:norm}
\begin{tabular}{llllll}
\toprule
           & \textbf{Test acc.}       & \textbf{Train acc.}      & \multicolumn{3}{l}{\textbf{Alignment cosine similarity}} \\
           &                 &                 & 1st layer      & 2nd layer      & 3rd layer     \\
           \cmidrule(r){2-6} 
norm      & 46.46\% (0.35)  & 86.62\% (0.31)  & 0.79 (0.08)    & 0.87 (0.10)    & 0.64 (0.07)   \\
no norm      & 43.46\% (0.27)  & 96.43\% (0.22)  & 0.64 (0.08)    & 0.69 (0.08)    & 0.65 (0.06)   \\
norm+BN & 45.73\% (0.53) & 84.52\% (0.30) & 0.62 (0.21)    & 0.43 (0.13)    & 0.56 (0.14) \\
norm+$\lvert x \rvert$ & 50.21\% (0.39) & 62.17\% (0.35) & 0.79 (0.02)    & 0.64 (0.04)    & 0.19 (0.02) \\
norm+ReLU & 40.24\% (2.16) & 41.68\% (0.40) & 0.38 (0.33)    & 0.37 (0.33)    & 0.06 (0.07) \\
norm+lReLU(-0.5) & \textbf{52.94}\% (0.38) & 68.67\% (1.73) & 0.84 (0.02)    & 0.72 (0.02)    & 0.22 (0.02) \\
norm+lReLU(0.01) & 37.88\% (4.34) & 41.46\% (0.69) & 0.42 (0.35)    & 0.40 (0.34)    & 0.07 (0.07) \\
norm+DO(0.5) & 47.90\% (0.17) & 54.23\% (0.15) & 0.72 (0.07)    & 0.73 (0.08)    & 0.43 (0.06) \\
norm+DO(0.1) & 49.50\% (0.34) & 69.97\% (0.18) & 0.79 (0.07)    & 0.83 (0.05)    & 0.54 (0.03) \\
\bottomrule
\end{tabular}
\end{table}

\paragraph{Batch Normalization} Batch Normalization (BN) \cite{ioffe2015} is an essential component of modern architectures. It has repeatedly proven its effectiveness, even though there has been much discussion about the exact underlying mechanisms \cite{santurkar2018,kohler2019}. In \cite{liao2016} it was reported as being essential to good performance with FA, while \cite{nokland2016} and \cite{lillicrap2016} did not use it. 

We argue that for DFA, BN is not critical like it is for FA. For fully-connected networks (Table \ref{tab:norm}), we confirm that BN is not necessary for good performance. In our specific test case, BN gives slightly worse performance and penalizes the average alignment angle significantly, also increasing its dispersion around the mean. For convolutional networks (Table \ref{tab:conv}), adding BN leads to a catastrophic loss of performance. As DFA is not actually training convolutional layers (see section \ref{sec:conv}), it is hard to draw a systematic conclusion regarding BN and DFA for such architectures.  

\paragraph{Dropout} Dropout \cite{srivastava2014} is a widely used technique to reduce overfitting in deep networks. Intuitively, DFA appears less sensitive to this issue, as it does not follow a precise gradient. Yet, our experiments in Table \ref{tab:norm} show that it still occurs. Accordingly, it's no surprise to see dropout enhancing performance, by limiting overfitting -- bringing the test and train accuracies closer together. 

However, dropout simultaneously reduces alignment: because it zeroes neurons randomly, it makes it harder for the feedforward weights to properly align. By reducing the dropout rate to 0.1 instead of the recommended 0.5, we can still benefit from its regularizing affect, while reducing its negative influence on DFA. Accordingly, while we recommend the use of dropout for DFA, it should be employed with lower rates than with BP. Moreover, if possible, other methods should be considered to mitigate overfitting, such as data augmentation. For convolutional architectures (Table \ref{tab:conv}), as the tendency to overfit is more limited, there are no benefits to using dropout. The penalty on alignment is simply too important and not worth the additional regularization.

\paragraph{Activation functions} In Table \ref{tab:norm}, we observe that ReLU is heavily penalizing both in performance and alignment for DFA, with a large dispersion of angle values. In layers close to the output, we observe angles close to 90\degree, denoting a complete failure to align. Also, the same analysis is valid for LeakyReLU (lReLU) \cite{maas2013} with negative slope $0.01$. The experiments with $\tanh{(x)}$ suffered of vanishing gradients problems. Indeed, synthetic gradients of DFA can still vanish either when the error signal becomes zero, or when the derivative of the activation function does.\\
Interestingly, a less common choice of activation function such as $\lvert x \rvert$ yielded high alignment values and good accuracy. Motivated by this observation, we tried lReLU with a negative slope of $-0.5$, to mimic a "weak" version of the absolute value. This function yielded the best results in our experiments. 
These observations point to a need of rethinking activation functions designed for BP in the context of training with DFA.

\section{Convolutional networks and the bottleneck effect}\label{sec:conv}
While simple convolutional architectures can be trained with FA or DFA, the gap in performance compared to BP is far more important. This remains unexplained so far, and has prevented DFA from scaling to harder computer vision tasks like ImageNet \cite{bartunov2018}. On the other hand, previous research has briefly mentioned the inability of FA and DFA to train architectures with narrow layers. We hypothesize that these two phenomena are one and the same.  

\subsection{DFA fails to train the convolutional part of CNNs}

\begin{table}[t]
\centering
\caption{Top-1 accuracy of VGG-16 on CIFAR-100 and ImageNet with BP and DFA. CIFAR-100 images were scaled up to (256x256). Learning rate was initially set to $10^{-2}$ and was decreased by a factor of 10 if the loss stagnated for more than 5 epochs on a separate validation test. Data augmentation was applied in the form of a random crop to size (224x224) and random horizontal flips. Results reported with an asterisk come from \cite{simonyan2014very}.}
\label{tab:cnndfa}
\begin{tabular}{lll}
\toprule
                   & BP       & DFA   \\
                   \cmidrule{2-3}
CIFAR-100 (scaled) & 60 \%    & 1\%   \\
ImageNet           & 75.2 \%* & 0.7\% \\
\bottomrule
\end{tabular}
\end{table}

\begin{figure}[b]
  \centering
  \includegraphics[width=0.8\linewidth]{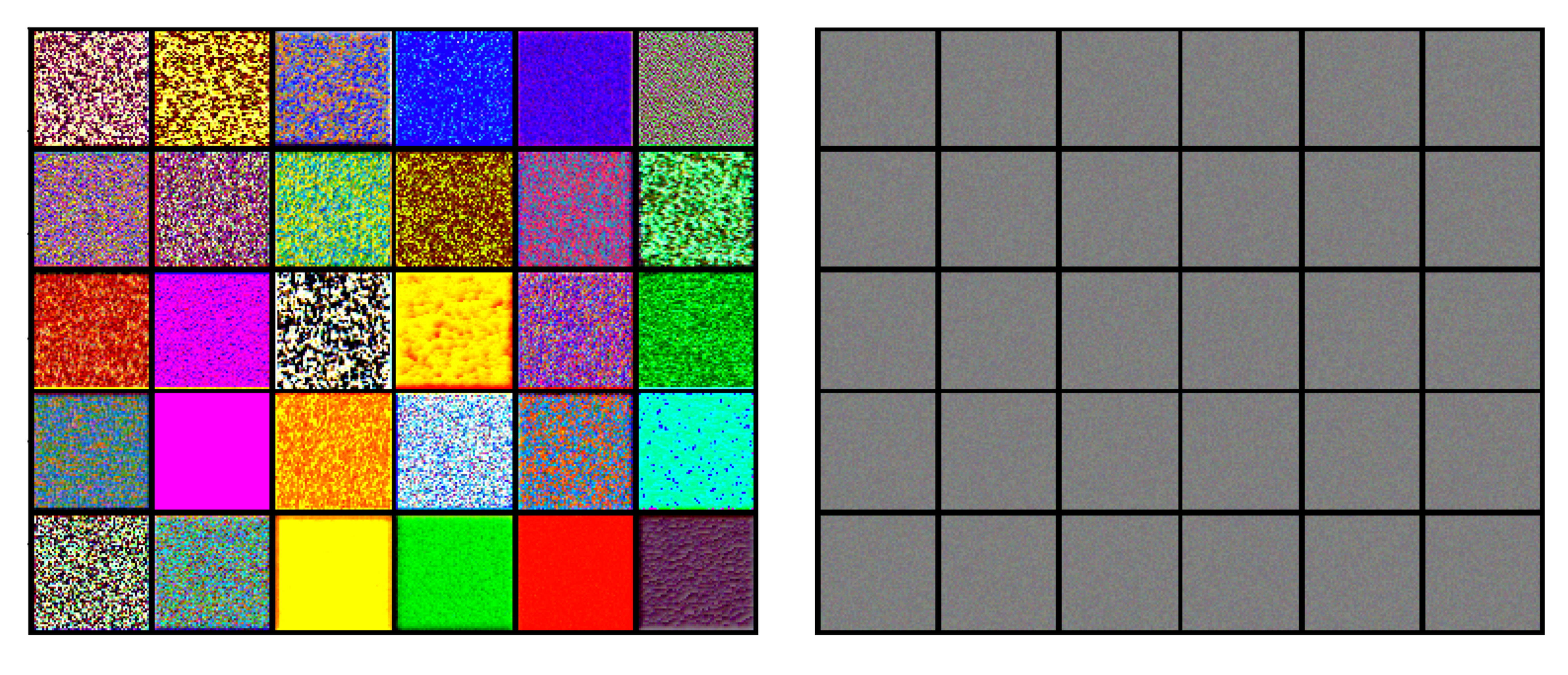}
  \caption{Visualization of the filters in the second convolutional layer of a VGG-16 architecture trained with BP (left) and DFA (right). VGG-16 is trained according to the procedure described in Table \ref{tab:cnndfa}. Visualizations are generated by finding images that maximize activations of a given filter. No regularization was used.} 
  \label{fig:filters}
\end{figure}

In \cite{nokland2016}, it was observed that DFA was significantly worse than BP at training convolutional architectures -- more so than for fully-connected networks. Nevertheless, this did not lead to further investigations, as CNNs trained with DFA were still outperforming their fully-connected counterparts as expected. In \cite{bartunov2018}, DFA was not scaled to classic convolutional networks like VGG-16 due to memory concerns. Using our new unique feedback matrix method, we are able to train VGG-16 with DFA on both CIFAR-100 and ImageNet. 

Results are reported in Table \ref{tab:cnndfa}: it is obvious that DFA completely fails to train these large architectures. To better understand theses results, we generated visualizations of the learned convolutional filters (Figure \ref{fig:filters}). The objective was not derive precise visualizations, but rather to be able to roughly compare the filters learned by BP and DFA. The results are quite clear: while BP creates meaningful filters, potentially extracting complex features, the filters learned by DFA are completely random.

Decent results for CNNs trained with DFA are reported by \cite{nokland2016} and ourselves in Table \ref{tab:conv} because they are on simple tasks. Most of the performance can then be attributed to the classifier. Performance is higher than a simpler fully-connected network, because random convolutional filters can still extract useful features \cite{saxe2011}. 

\subsection{The bottleneck effect: constrained degrees of freedom prevent alignment}

\begin{table}[t]
\centering
\caption{Performance of a CNN with 3 convolutional layers and a one layer classifier on CIFAR-10 with different practices. Learning rate is set to $5  \cdot 10^{-4}$ for SGD with no momentum. Weights are initialized using He initialization. Non-linearities are tanh. Accuracies reported are the average of 10 random runs. Statistics for alignment angles are computed on a batch of 128 samples from the best run of the 10. Values in parentheses are standard deviations.}
\label{tab:conv}
\begin{tabular}{llllll}
\toprule
           & \textbf{Test acc.}      & \multicolumn{4}{l}{\textbf{Alignment cosine similarity}}       \\
           &                                & conv. 1     & conv. 2     & conv. 3     & class. 4    \\
           \cmidrule(r){2-6} 
norm      & \textbf{62.65}\% (0.65)  & 0.02 (0.02) & 0.15 (0.04) & 0.59 (0.06) & 1.0 (0.00) \\
no norm      & 57.90\% (0.82) & 0.00 (0.08) & 0.09 (0.12) & 0.62 (0.12) & 1.00 (0.00) \\
norm+BN & 48.50 \% (1.52) & 0.01 (0.06) & 0.00 (0.07) & 0.00 (0.11) & 0.14 (0.20) \\
norm+DO(0.5) & 57.95 \% (1.02) & 0.01 (0.02) & 0.09 (0.04) & 0.30 (0.11) & 0.96 (0.05) \\
norm+DO(0.1) & 61.31 \% (0.78) & 0.02 (0.01) & 0.12 (0.04) & 0.44 (0.12) & 0.98 (0.04) \\
\bottomrule
\end{tabular}
\end{table}

\paragraph{Bottlenecks and alignment} Narrower layers lack the degrees of freedom necessary to allow for feedforward weights to both learn the task at hand, and align with the random feedback weights.To better quantify this effect, we devise an experiment in which the second layer of a three layers fully-connected classification network is artificially bottlenecked. 

Doing so by simply reducing the number of neurons in the layer would also hamper the forward pass, by preventing information flow. Instead, we keep its size constant, but zero out a certain percentage of the elements of its gradients. By doing so, a fixed number of weights will remain constant throughout training, and will not be able to contribute to alignment -- but will still let information flow through in the forward pass. The parts of the gradient that are zeroed out are selected at initialization time with a random mask, and that mask is kept the same for all of the training.

\begin{figure}[h]
  \centering
  \includegraphics[width=\linewidth]{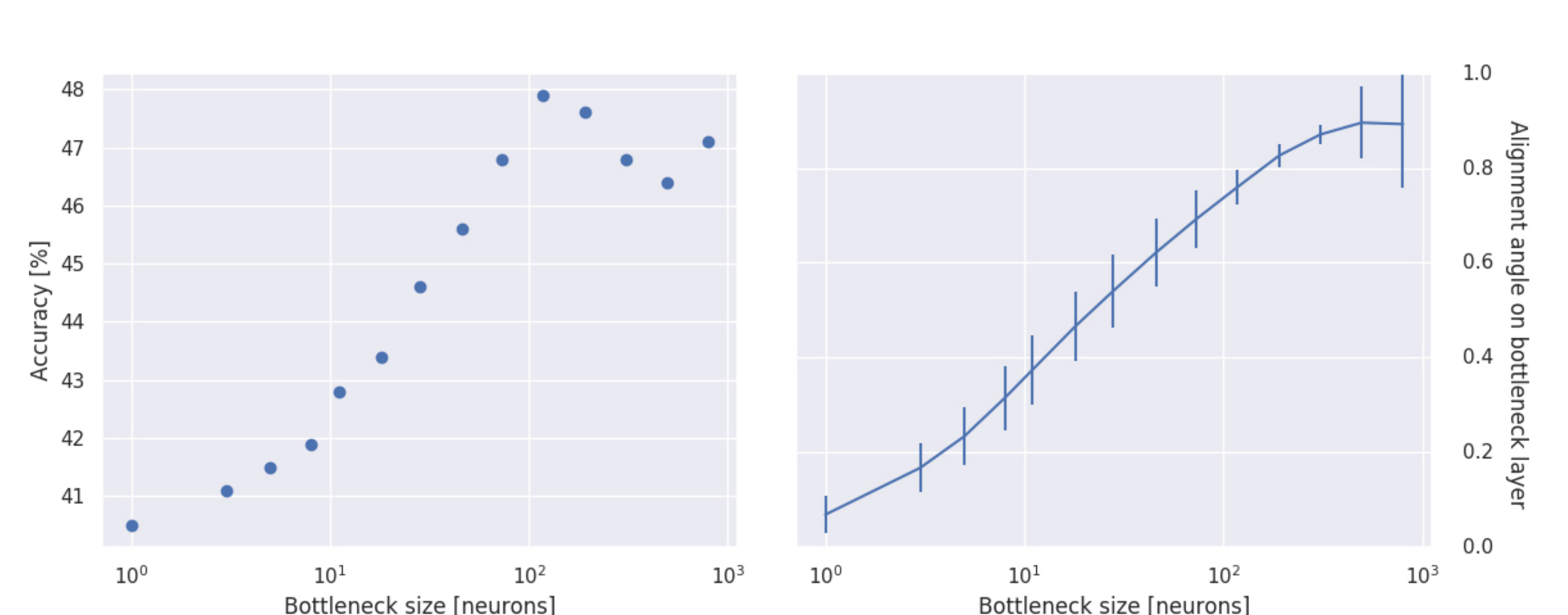}
  \caption{Accuracy and alignment for various bottlenecks. The size of the bottleneck is the number of neurons with non zeroed out gradients. Architecture is a 3 layers of 800 neurons network trained on CIFAR-10. The second layer is bottlenecked. Learning rate is set to $5  \cdot 10^{-3}$ for SGD with no momentum. Weights are initialized using He initialization and non-linearities are tanh. Statistic for alignment are computed on a batch of 128 samples.}
  \label{fig:bottleneck}
\end{figure}

The results obtained are reported in Figure \ref{fig:bottleneck}. We observe an initial phase, where an increase in the number of neurons able to align leads to an increase in accuracy and in alignment in the bottleneck. Eventually, this settles out, and performance and alignment remains more or less constant. 

The high base performance at the beginning can be explained by the relative simplicity of the task: the last layer is still able to get make decent classification guesses as information can freely flow through the bottlenecked layer in our scheme. We expect the threshold for performance to remain constant -- here located at around 100 neurons -- to change depending on the difficulty of the task.

\paragraph{Convolutions as bottlenecks} 
Convolutional layers benefit of fewer degrees of freedom than fully connected ones as they must obey to a certain structure. Accordingly, convolutions are not properly trained by DFA because they create bottlenecks in the network. This is corroborated by results in Table \ref{tab:conv}: alignment angles in deep convolutional layers are zeroes. This means the update derived by DFA are orthogonal to the ones of BP; they are essentially random. 

\section{Conclusion and outlooks}
A thorough and principled analysis of the effects of dropout and batch normalization for the training of deep neural networks with DFA has shown that the recommendations that apply for BP don't translate easily to this synthetic gradient method. Similarly, the design and choice of activation functions in this context needs adaptation and will be the subject of future work. These new insights have been made possible by direct measurement of the alignment angles, allowing for a better comprehension of how different practices affect the learning mechanics of DFA.\\
We have more precisely characterized a bottleneck effect, that prevents learning with DFA in narrower layers. Due to their structured nature, convolutional layers suffer from this effect, preventing DFA from properly training them -- as is verified by measurement of the alignment angles. For DFA to be scaled to harder computer vision tasks, this issue needs to be tackled, either by a change in the formulation of DFA, or adaptations of convolutional layers to it.

In the future we plan to expand our analysis to different types of structured layers, like \cite{yang2015} and \cite{moczulski2016}, and delve deeper into the interactions between DFA and convolutional layers. We also see paths to further computational optimizations in this method, for example by using a Fastfood transform \cite{le2013} in place of the dense random matrix multiplication in the backward pass.

\bibliographystyle{unsrt}
\bibliography{bibliography}

\end{document}